%% file: main.tex
\documentclass[runningheads]{llncs}

\usepackage[mobile]{eccv}

\usepackage{eccvabbrv}

\usepackage{graphicx}
\usepackage{booktabs}
\usepackage{amsmath}
\usepackage{bm}
\usepackage{bbm}
\usepackage{mathtools}
\usepackage{lipsum}
\usepackage{xcolor}
\usepackage[dvipsnames]{xcolor}
\usepackage[table]{xcolor}
\usepackage{bbding}
\usepackage{subcaption}
\usepackage{ulem}
\usepackage{wrapfig}
\DeclarePairedDelimiterX{\js}[2]{}{}{%
  #1\;\delimsize\|\;#2%
}

\usepackage[accsupp]{axessibility}  

\usepackage[pagebackref,breaklinks,colorlinks,citecolor=eccvblue]{hyperref}

\usepackage{orcidlink}

\begin{document}

\title{UNIC: Neural Garment Deformation Field for Real-Time Clothed Character Animation} 


\author{
Chengfeng Zhao\inst{1} \and
Junbo Qi\inst{2} \and
Yulou Liu\inst{1} \and
Zhiyang Dou\inst{3} \and
Minchen Li\inst{4} \and \\
Taku Komura\inst{5} \and
Ziwei Liu\inst{6} \and
Wenping Wang\inst{7} \and
Yuan Liu\inst{1}
}


\institute{$^{1}$HKUST, $^{2}$Waseda, $^{3}$MIT, $^{4}$CMU, $^{5}$HKU, $^{6}$NTU, $^{7}$TAMU\\
\href{https://igl-hkust.github.io/UNIC/}{https://igl-hkust.github.io/UNIC/}
}

\maketitle

\input{sections/0-abstract}
\input{sections/1-introduction}
\input{sections/2-rw}
\input{sections/3-method}
\input{sections/4-exp}
\input{sections/5-conclusion}

\newpage

%
%
\bibliographystyle{splncs04}
\bibliography{main}

\end{document}

%% file: sections/0-abstract.tex
\begin{abstract}
Simulating physically realistic garment deformations is an essential task for virtual immersive experience, which is often achieved by physics simulation methods. However, these methods are typically time-consuming, computationally demanding, and require costly hardware, which is not suitable for real-time applications. Recent learning-based methods tried to resolve this problem by training graph neural networks to learn the garment deformation on vertices, which, however, fail to capture the intricate deformation of complex garment meshes with complex topologies. In this paper, we introduce a novel neural deformation field-based method, named UNIC, to animate the garments of an avatar in real time, given the motion sequences. Our key idea is to learn the instance-specific neural deformation field to animate the garment meshes. Such an instance-specific learning scheme does not require UNIC to generalize to new garments but only to new motion sequences, which greatly reduces the difficulty in training and improves the deformation quality. Moreover, neural deformation fields map the 3D points to their deformation offsets, which not only avoids handling topologies of the complex garments but also injects a natural smoothness constraint in the deformation learning. Extensive experiments have been conducted on various kinds of garment meshes to demonstrate the effectiveness and efficiency of UNIC over baseline methods, making it potentially practical and useful in real-world interactive applications like video games.
\keywords{real-time clothed character animation \and neural field \and neural garment simulation}
\end{abstract}

%% file: sections/1-introduction.tex
\section{Introduction}
\label{sec:intro}

Efficiently simulating physically realistic garment deformations is an essential task in computer graphics, which benefits various downstream tasks, including gaming, fabrication, and the metaverse. However, efficiently simulating complex garment deformation remains challenging due to the high computational cost of physics-based methods, even when utilizing professional software~\cite{md,clo3d,style3d} with GPU acceleration. Moreover, recent advancements in differentiable~\cite{liang2019differentiable,hu2020difftaichi,li2022diffcloth} and neural simulators~\cite{kair2024neuralclothsim} show huge potential in cooperating with deep learning frameworks through a gradient-available physics-based solver. However, they are still unable to support real-time and interactive applications due to numerous optimization iterations for satisfying the physical constraints of garments.

\begin{figure*}[t!]
  \includegraphics[width=\textwidth]{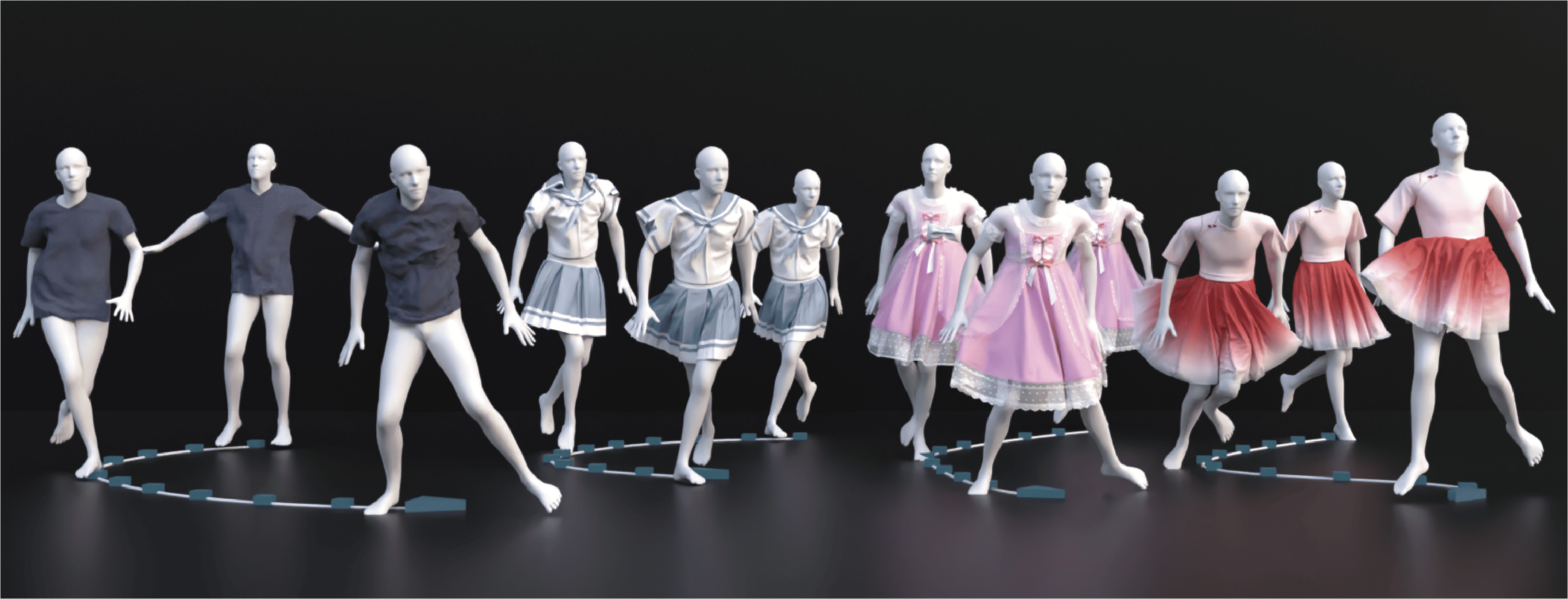}
  \vspace{-20pt}
  \caption{
  UNIC enables real-time and high-quality physically realistic deformation of complex garments with arbitrary topologies to follow arbitrary unseen character motions, which benefits real-time character animation applications like video games.
  }
\label{fig:teaser}
\vspace{-15pt}
\end{figure*}

In recent years, data-driven approaches~\cite{lahner2018deepwrinkles,gundogdu2019garnet,santesteban2019learning,patel2020tailornet,santesteban2022snug,zhang2022motion,grigorev2023hood,grigorev2024contourcraft,cao2023efficient} have emerged to directly learn cloth deformation using neural networks. Among these, Graph Neural Network (GNN) based methods~\cite{patel2020tailornet,santesteban2022snug,grigorev2023hood,xu2018powerful} are particularly prevalent. By leveraging simulated datasets~\cite{AMASS:ICCV:2019,santesteban2019learning}, these models achieve a certain degree of generalization to unseen poses and garment shapes. However, this generalization is inherently restricted to structurally simple garments within a strict style domain. Because GNNs fundamentally rely on propagating and aggregating features along predefined graph edges, their representation power is heavily constrained by the specific topologies seen during training.

Consequently, when confronted with the arbitrary topologies and intricate sewing patterns of unseen garments, these methods fail to extract robust graph features. Crucially, the ability to animate only simple garments offers limited practical value. Real-world applications, such as digital fashion, gaming, and the metaverse, intrinsically demand the simulation of highly complex and diverse clothing. How to effectively extend these neural approaches to accommodate arbitrary, complex topologies while maintaining high-fidelity deformation quality remains an open and unresolved challenge. As a result, the inability to bridge the gap between simple topological memorization and complex real-world garment diversity significantly limits the practical deployment of these existing methods.

To bridge the gap between academic research and practical deployment, we rethink the setting of neural garment simulation. Instead of pursuing the overly ambitious but practically flawed goal of generalizing across arbitrary garment styles and body shapes, which inevitably fails on complex clothing, we propose a paradigm shift towards an instance-specific setting. In this paper, we introduce UNIC, a novel framework designed to be truly functional for real-world applications. By constraining the generalization requirement strictly to unseen motions for a specific character and garment, UNIC drastically simplifies the learning complexity. This deliberate trade-off allows our method to guarantee high-fidelity, physically realistic deformations even for highly complex garments, making it practically viable for downstream tasks.

\begin{wrapfigure}{r}{0.42\textwidth}
\vspace{-30pt}
  \begin{center}
    \includegraphics[width=0.4\textwidth]{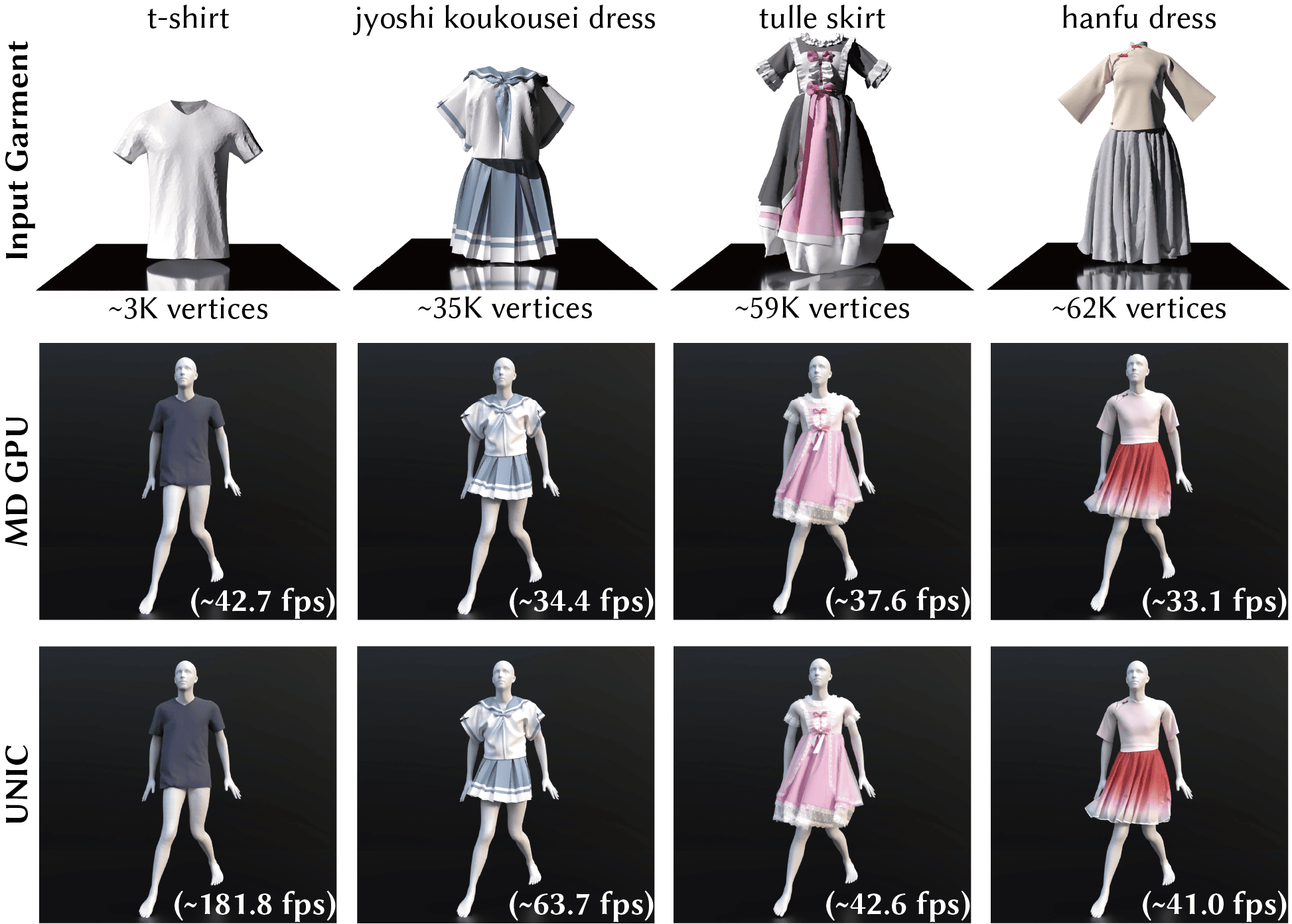}
  \end{center}
  \vspace{-10pt}
  \caption{\textbf{Quality and efficiency comparison to the ``gold standard''.} Given garments of different topology complexity and geometry density (top row), UNIC consistently outperforms GPU-accelerated professional software~\cite{md} in efficiency, while also achieving comparable simulation quality (middle and bottom row). The inference speed is measured on a single NVIDIA RTX 3090 GPU.}
\label{fig:intro}
\vspace{-20pt}
\end{wrapfigure}

However, designing an effective architecture even for this focused setting is non-trivial. As previously discussed, typical Graph Neural Network (GNN) architectures are heavily constrained by predefined training topologies, rendering them incapable of handling the arbitrary and intricate geometries of complex real-world clothing. To overcome this structural limitation, we formulate garment deformation as the learning of a continuous neural field. Unlike previous works~\cite{gundogdu2019garnet,patel2020tailornet,zhang2022motion,grigorev2023hood} that predict per-vertex offsets relative to the character body, UNIC uses a Multi-Layer Perceptron (MLP) to map arbitrary 3D spatial coordinates to deformation offsets. Inspired by recent advances in neural fields~\cite{mildenhall2021nerf,wang2021neus}, this representation offers an inherent spatial smoothness prior and is completely agnostic to garment topology or resolution, granting us the extreme flexibility needed to model complex sewing patterns and intricate wrinkles.

Another critical challenge in real-time simulation lies in ensuring that this continuous deformation field generalizes robustly to unseen motions and remains stable over long-term autoregressive execution (e.g., continuous gameplay). Directly learning a mapping from pose variations to garment deformations often leads to poor generalization on novel motions and severe error accumulation (drift) over time. To address this, we propose encoding the previous and current character poses into a discrete, categorical motion latent space~\cite{starke2024categorical}. By mapping the character's motion state into these categorical features before querying the neural deformation field, UNIC effectively regularizes the motion representation. This strategic design not only ensures robust generalization to unseen motions but also guarantees that autoregressive predictions remain stable and drift-free, even during extended runtimes.

Furthermore, to ensure visual plausibility in highly dynamic interactive scenarios, we introduce a lightweight Intersection Handling module into the UNIC framework. This module effectively resolves unwanted penetrations between the complex garments and the character's body, ensuring collision-free and physically credible results.

To evaluate the effectiveness of UNIC, we conduct extensive experiments using the standard SMPL character~\cite{SMPL2015} paired with a wide variety of garments, ranging from simple T-shirts to highly complex dresses simulated via professional physics software~\cite{md,clo3d}. The results demonstrate that UNIC significantly outperforms existing baseline methods~\cite{patel2020tailornet,santesteban2022snug,zhang2022motion,grigorev2023hood} in both deformation quality and computational efficiency, particularly on intricate clothing. Finally, we provide real-time interactive demos deployed in the Unity Engine, concretely validating that our method is not merely a theoretical exploration, but a highly practical solution ready for real-time applications.

%% file: sections/2-rw.tex
\section{Related Work}
\label{sec:rw}
Realistic and efficient garment simulation is a core topic in gaming, animation and AR/VR. Since our work proposes to utilize neural networks to simulate garment deformations for real-time animations, we first review traditional physics-based methods, and then discuss about learning-based methods as an efficient alternative. Finally we talk about techniques for real-time character and cloth animation applications.

\vspace{-2mm}
\paragraph{\bf Physics-based Garment Simulation.} Physics-based simulation methods have been extensively studied in past decades, evolving in various directions such as fabric modeling \cite{terzopoulos1987elastically,provot1995deformation,grinspun2003discrete,bhat2003estimating,choi2005stable,nealen2006physically,volino2009simple,wang2011data,miguel2012data,chen2023multi}, collision handling \cite{bridson2002robust,harmon2009asynchronous,tang2018cloth,li2020codimensional}, subspace methods \cite{hahn2014subspace}, and implicit time integration solvers \cite{thomaszewski2008asynchronous,baraff2023large,li2023subspace}. Based on these techniques, industrial-level products \cite{md,clo3d,style3d} have achieved great success in fashion design, virtual try-on and animation. Recently, researchers have paid more attention to developing differentiable simulators \cite{liang2019differentiable,hu2020difftaichi,li2022diffcloth}, which can coordinate with gradient-based optimization to improve fidelity and extend to garment reconstruction and tracking. Shifting from traditional continuum models, yarn-level models \cite{kaldor2008simulating,cirio2014yarn} offer more detailed visual outcomes in two-way coupling simulations that integrate garments with dynamic avatars \cite{montes2020computational,romero2020modeling}. Despite significant progress, physics-based garment simulation remains computationally expensive, which poses ongoing challenges for real-time applications.

\vspace{-2mm}
\paragraph{\bf Neural Garment Deformation Learning.} Learning-based methods emerged as a solution to address the computational inefficiencies of physics-based simulation, leveraging statistical models to predict garment deformations in a data-driven manner. Taking the philosophy of PSD \cite{lewis2023pose} and parametric body models \cite{SMPL2015,MANO:SIGGRAPHASIA:2017,SMPLX2019}, earlier studies \cite{guan2012drape,wang2018learning,santesteban2019learning,ma2020learning,patel2020tailornet} model garment deformations as a function of body pose and shape parameters. However, these methods rely on proximal skinning weights, which tend to overlook intricate wrinkles and realistic fabric deformations, especially on complex garment styles. Alternatively, deep learning based approaches learn garment feature in Euclidean \cite{wang2019learning,pfaff2020learning,sanchez2020learning,santesteban2021self,bertiche2021deepsd,fortunato2022multiscale,pan2022predicting,shao2023towards,grigorev2023hood,liu2026pb4u,guo2025garmentx,sun2025tailor,li2025etch,chen2024single,wang2025garmentcrafter,li2025single,li2025dress123,can2026image2garment,li2025garmentdreamer} and UV space \cite{holden2019subspace,zhang2022motion,vidaurre2025diffusedwrinkles}, thereby facilitating multi-layer garment simulation \cite{li2024isp,grigorev2024contourcraft} and enabling flexible geometry editing \cite{su2022deepcloth}. In a separate line of research, self-supervised work \cite{bertiche2020pbns,santesteban2022snug} formulate intrinsic physical constraints of fabric dynamics as loss terms, avoiding the dependence on expansive high-quality datasets \cite{wang2011data,narain2012adaptive,narain2013folding,pfaff2014adaptive,bertiche2020cloth3d,zou2023cloth4d,wang20244d,li2025garmagenet}. Although learning-based methods achieved impressive results with significant efficiency gains, most of them adhere to the paradigm that decomposes garment dynamics into deformations in canonical space and global posing through skinning \cite{lahner2018deepwrinkles,yang2018analyzing,buffet2019implicit,vidaurre2020fully,corona2021smplicit,chen2021snarf,ma2021power,lin2022learning,santesteban2022snug,santesteban2022ulnef,li2023swingar,wang2024towards}, which struggles with handling modern fashion designs characterized by diverse sewing patterns. Neural fields for generic meshes~\cite{aigerman2022neural} and garment dynamics~\cite{li2024neural} are developed well, but miss the awareness of character motion. To solve these limitations, we propose to utilize implicit representation of garment vertices informed with character motion, and learn instance-specific neural fields for arbitrary garment styles on physically simulated data.

\vspace{-2mm}
\paragraph{\bf Real-time Character and Cloth Animation.} Traditional motion matching algorithm \cite{clavet2016motion} is developed to realize real-time character animation, which searches and stitches nearest neighbor within a virtual motion graph constructed by pre-recorded motion clips. To improve the issues of low diversity and memory inefficiency, learning-based methods \cite{bergamin2019drecon,holden2020learned,li2023example} proposed to compress the whole motion graph into neural networks, and learn states transitioning probability distribution for sampling next-step character motion instead of searching. Alternative to applying motion matching in Euclidean space, some approaches learn cyclic phase function from motion capture datasets as character motion representation \cite{holden2017phase,starke2022deepphase}, achieving better similarity measure. Extending from single character animation, various methods apply the similar matching strategy on quadruped character motion control \cite{zhang2018mode}, character-scene interactions \cite{starke2019neural}, multi-contact character movements \cite{starke2020local}, and embodiment in VR gaming \cite{starke2024categorical}, demonstrating the efficacy of representing continuous motion space as neural graphs. Similarly, real-time cloth animation is a long-standing and unsolved problem for computer graphics. Early methods attempted to apply physical simulation on low-resolution garments \cite{feng2010deformation,gillette2015real} or take sensitivity-optimized rigging \cite{xu2014sensitivity} to add realistic dynamic wrinkles for real-time but coarse animation. However, we found that real-time matching algorithms in character animation can be applied to cloth, and more fundamentally, clothed character animation, which has not been extensively explored yet. In this paper, we take the motion matching philosophy to construct neural graphs for clothes, and animate the clothed character with realistic clothing deformations in real time.

%% file: sections/3-method.tex
\section{Method}
\label{sec:method1}

\begin{figure*}[t!]
  \centering
  \includegraphics[width=\linewidth]{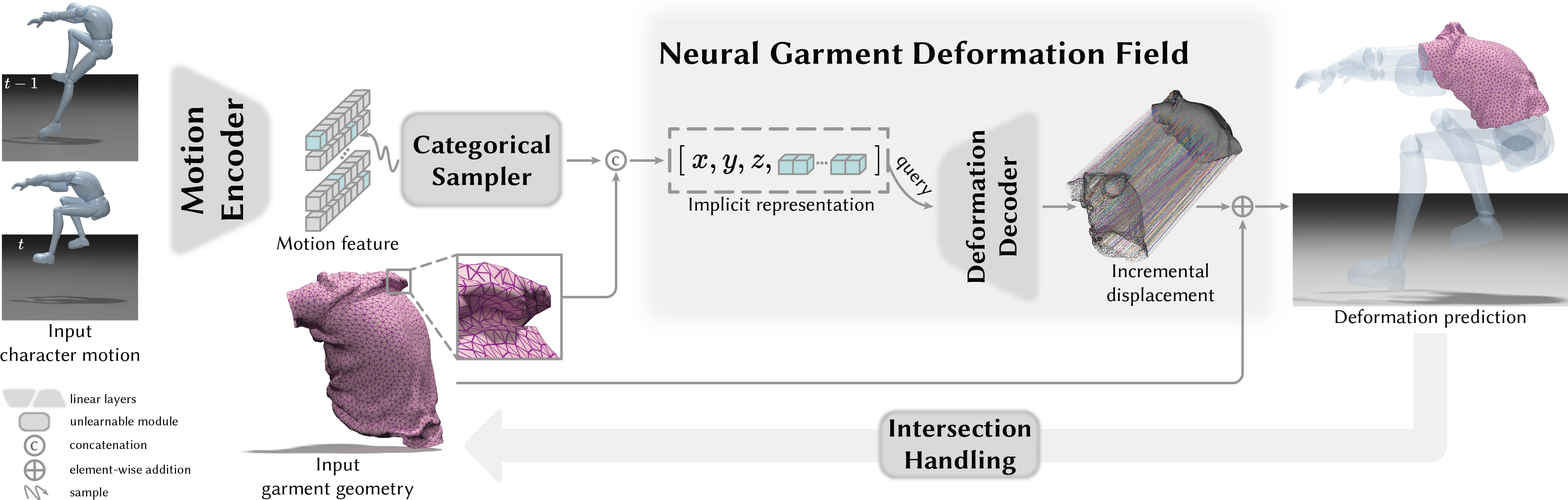}
  \vspace{-15pt}
  \captionof{figure}{\textbf{Overview of our training pipeline.} UNIC learns an instance-specific neural deformation field that deforms arbitrarily complex garments to follow unseen character poses in real time. We first encode the character poses of two consecutive frames into a compact latent space. Then, we categorically sample a latent vector from the learned space and concatenate it with garment vertex coordinates. After that, we feed the concatenation into an MLP-based deformation decoder to predict deformation offsets for all vertices. Finally, post-process intersection handling is applied to avoid the intersection between the avatar mesh and the deformed garment mesh.}
\label{fig:train_pipeline}
\vspace{-15pt}
\end{figure*}

Given two consecutive frames of character poses and the garment geometry in the previous frame, our goal is to predict physically realistic garment deformation in the current frame. An overview of our method is depicted in Fig.~\ref{fig:train_pipeline}. We first introduce the categorical character motion encoder in Sec.~\ref{sec:encoder}, which learns a compact and expressive latent space capturing transitions between motion states. 
Subsequently, in Sec.~\ref{sec:neural_field}, we propose the instance-specific neural deformation fields to learn the deformation of garments, which can handle arbitrarily complex garment topologies with high inference efficiency. Moreover, to avoid occasional intersections during inference, we devise a differentiable intersection handling module to ``drag'' intersected garment vertices outside the character body surface with a relaxation buffer (see Sec. \ref{sec:intersection_handling}).

\subsection{Categorical Character Motion Encoder}
\label{sec:encoder}
Garment deformations follow complex character motion state transition in the temporal dimension, rather than a single pose. Therefore, instead of directly mapping character poses onto garment deformation space through proximal skinning weights~\cite{lahner2018deepwrinkles,gundogdu2019garnet,patel2020tailornet,santesteban2019learning,zhang2022motion,santesteban2022snug,grigorev2023hood}, we encode consecutive two-frame character poses into a compact low-dimensional latent space, which learns the transitioning feature between two consecutive motion states. 

\vspace{-2mm}
\paragraph{\bf Motivation of our design.} Learning such a motion manifold for informing garment deformations is non-trivial because character motions within a small context window can produce ambiguities. Served as conditions or part of the garment state representation, inexpressive latent features of diverse character motions can result in cumulative errors in the autoregressive inference of garment deformations. Inspired by CM~\cite{starke2024categorical}, we learn a categorical latent motion space alternative to the continuous motion manifold, which allows sampling from a valid set of vector-quantized latent codes to mimic the transition between exemplar character states~\cite{starke2019neural}. Unlike continuous motion space that may suffer from error accumulation with drifting, the discrete categorical motion space helps alleviate the distribution shift problem during autoregressive inference in the following neural deformation field.

\vspace{-2mm}
\paragraph{\bf Motion Representation.} The input to the categorical character motion encoder are two character poses in two consecutive frames 
\begin{equation}
    \mathbf{x}_t=\left[\mathbf{c}_{t-1},\mathbf{c}_t\right]\in\mathbb{R}^{2D_c}.
\label{eq:input_x}
\end{equation}
The character motion at the $i$-th frame is defined as
\begin{equation}
    \mathbf{c}_{i} = \left[
        \mathbf{r},
        \dot{\mathbf{r}},
        \mathbf{\Phi} ,
        \dot{\mathbf{\Phi}},
        \mathbf{j}^r,
        \mathbf{j}^p,
        \mathbf{j}^v,
        \mathbf{p}
    \right] \in \mathbb{R}^{D_c},
\label{eq:motion_representation}
\end{equation}
which comprises global translation $\mathbf r\in\mathbb R^{3}$, global linear velocity $\dot{\mathbf{r}} \in\mathbb R^{3}$, global orientation $\mathbf \Phi\in\mathbb R^{6}$, and global angular velocity $\dot \Phi\in\mathbb R^{3}$. In addition, we represent rotation, position, and linear velocity of local joints as $\mathbf {j}^r\in \mathbb R^{6J}$, $\mathbf {j}^p\in\mathbb R^{3J}$, and $\mathbf {j}^v\in\mathbb R^{3J}$, with extra 4 binary labels $\mathbf p\in \mathbb R^{4}$ to annotate the contacts between ground and feet. In our implementation, we utilize standard SMPL~\cite{SMPL2015} skeleton with $J=23$ local joints and a 6D rotation convention~\cite{zhou2019continuity} for all rotational parts in the representation. Overall, the character motion representation has a dimension $D_c=295$.

\vspace{-2mm}
\paragraph{\bf Categorical Motion Feature Learning.} Given the input tuple $\mathbf{x}_t$, we extract the latent feature graph $\mathbf{Z}_t$ using the encoder $\mathcal{E}_{\theta}$
\begin{equation}
    \mathbf{Z}_t = \mathcal{E}_{\theta}(\mathbf{x}_t) \in \mathbb{R}^{D_h\times D_w}\text{.}
\label{eq:motion_feature}
\end{equation}
The learned feature graph consists of  $D_h$ channels of feature in $D_w$ dimensions. The encoder $\mathcal{E}_{\theta}$ is composed of 3 linear layers interleaved with exponential linear unit (ELU) activation function~\cite{clevert2015fast}. Then, we categorically sample a single latent code in each channel of the feature within $D_w$ dimension, resulting in a latent vector
\begin{equation}
    \tilde{\mathbf{z}}_t = \mathcal{G}(\mathbf{Z}_t) \in \mathbb{R}^{D_h}\text{,}
\label{eq:sampled_motion_feature}
\end{equation}
where $\mathcal{G}(\cdot)$ represents the categorical sampling operator with Gumbel-Softmax technique~\cite{jang2016categorical}. 

\subsection{Neural Garment Deformation Field}
\label{sec:neural_field}
Given the above-extracted motion latent vectors, we aim to deform arbitrary garments with realistic wrinkles in real time. 
Instead of using skinning weights to depict the garment deformations driven by motions~\cite{lahner2018deepwrinkles,santesteban2019learning,gundogdu2019garnet,patel2020tailornet,santesteban2022snug,zhang2022motion,grigorev2023hood}, we propose to represent the garment-motion state by concatenating the Euclidean coordinates of every garment vertex with the sampled motion feature. Taking such concatenated vectors as queries, our method learns a neural field directly on the garment deformation space, predicting the incremental displacements of each garment vertex concurrently.

\vspace{-2mm}
\paragraph{\bf Coordinate-based Garment-motion Representation.}
\begin{wrapfigure}{r}{0.42\textwidth}
\vspace{-50pt}
  \begin{center}
    \includegraphics[width=0.4\textwidth]{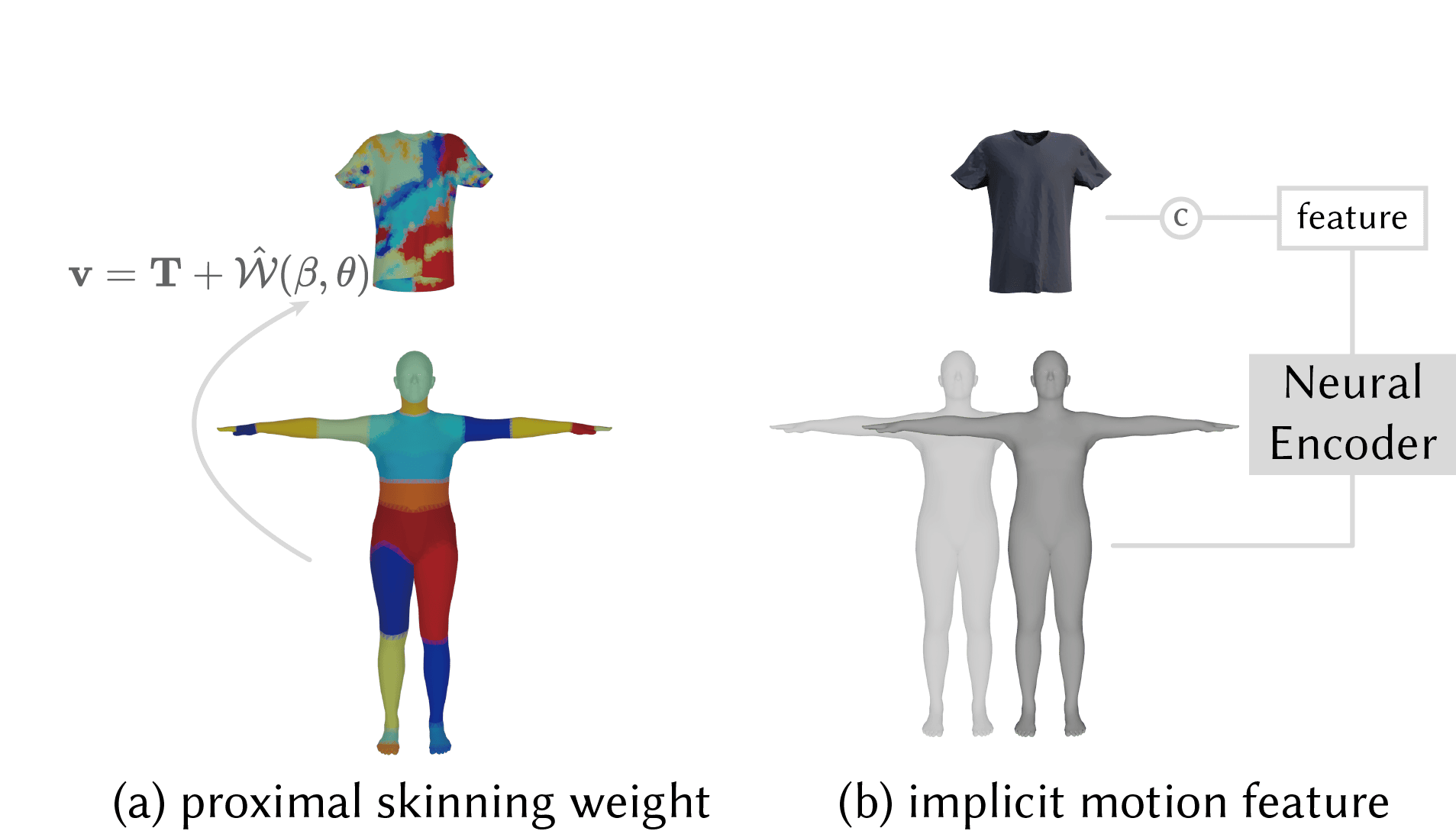}
  \end{center}
  \vspace{-20pt}
  \caption{\textbf{Different Garment-motion representations.} (a) Explicit representation determined by proximal skinning weights on top of parametric skeleton model~\cite{SMPL2015}. (b) Our representation combining garment vertex coordinates with motion latent features.}
\label{fig:representation}
\vspace{-20pt}
\end{wrapfigure}

To represent the interaction between garments and character motions, for a specific garment geometry $\mathbf{G}_t =\left\{\mathbf{v}_t^i \mid i\in\left\{1\dots V\right\}\right\}\in \mathbb{R}^{V\times 3}$ at timestamp $t$, we concatenate the 3D coordinate of each vertex $\mathbf{v}_t^i$ with the sampled motion latent vector $\tilde{\mathbf{z}}_t$ at the same time
\begin{equation}
    \tilde{\mathbf{v}}_t^i = \left[\mathbf{v}_t^i,\tilde{\mathbf{z}_t}\right] \in \mathbb{R}^{D_h+3},
\label{eq:implicit_coordinate}
\end{equation}
where the garment-motion state is represented by all the concatenated vectors of all the vertices of the garment in this equation.
As illustrated in Fig.~\ref{fig:representation}, existing methods represent garment geometries explicitly by linearly deforming them with skinning weights, which is inflexible and only considers one static pose. In contrast, we concatenate the motion feature vector with the garment vertex coordinates to represent the garment-motion states, which contain both local garment geometry and global character dynamics.

\vspace{-2mm}
\paragraph{\bf Autoregressive Deformation Decoder.} Given any implicit 3D coordinate $\tilde{\mathbf{v}}_{t-1}^i$ of $\hat{\mathbf{G}}_{t-1}$ at previous frame, our neural deformation field takes it as a query point, and learns to predict the incremental displacement $\Delta\hat{\mathbf{v}}_{t-1}^i$ by decoding
\begin{equation}
    \Delta\hat{\mathbf{v}}_{t-1}^i = \mathcal{D}_{\phi}\left(\tilde{\mathbf{v}}_{t-1}^i\right)\text{.}
\label{eq:delta_output}
\end{equation}
Consequently, all $V$ vertices of $\hat{\mathbf{G}}_{t-1}$ will be displaced by
\begin{equation}
    \hat{\mathbf{v}}_t^i = \hat{\mathbf{v}}_{t-1}^i + \Delta\hat{\mathbf{v}}_{t-1}^i\text{,}
\label{eq:final_output}
\end{equation}
and the deformed garment at the current frame is
\begin{equation}
    \hat{\mathbf{G}}_t =\left\{\hat{\mathbf{v}}_t^i \mid i\in\left\{1\dots V\right\}\right\}\in \mathbb{R}^{V\times 3}\text{.}
\end{equation}
In our implementation, the deformation MLP $\mathcal{D}_{\phi}$ has 8 layers with ReLU activation and skip connections~\cite{mildenhall2021nerf}.

\vspace{-2mm}
\paragraph{\bf Training Objectives.} UNIC is trained end-to-end using a single geometry loss term. To be specific, we train the neural deformation field $\mathcal{D}_{\phi}$ together with aforementioned motion encoder $\mathcal{E}_{\theta}$ using the supervision of ground-truth garment geometry $\mathbf{G}_t$
\begin{equation}
    \mathcal{L}\left(\theta,\phi,\mathbf{x}_t,\hat{\mathbf{G}}_{t-1}\right) = \lambda\frac{1}{V}\left\Vert\hat{\mathbf{G}}_t - \mathbf{G}_{t}\right\Vert_2^2\text{.}
\label{eq:output}
\end{equation}
The end-to-end training framework ensures the construction of motion manifold is informed by garment deformation learning, which helps maintain the critical motion transitions of the character that determine garment deformations.

\subsection{Intersection Handling}
\label{sec:intersection_handling}
\begin{wrapfigure}{r}{0.42\textwidth}
\vspace{-80pt}
  \begin{center}
    \includegraphics[width=0.32\textwidth]{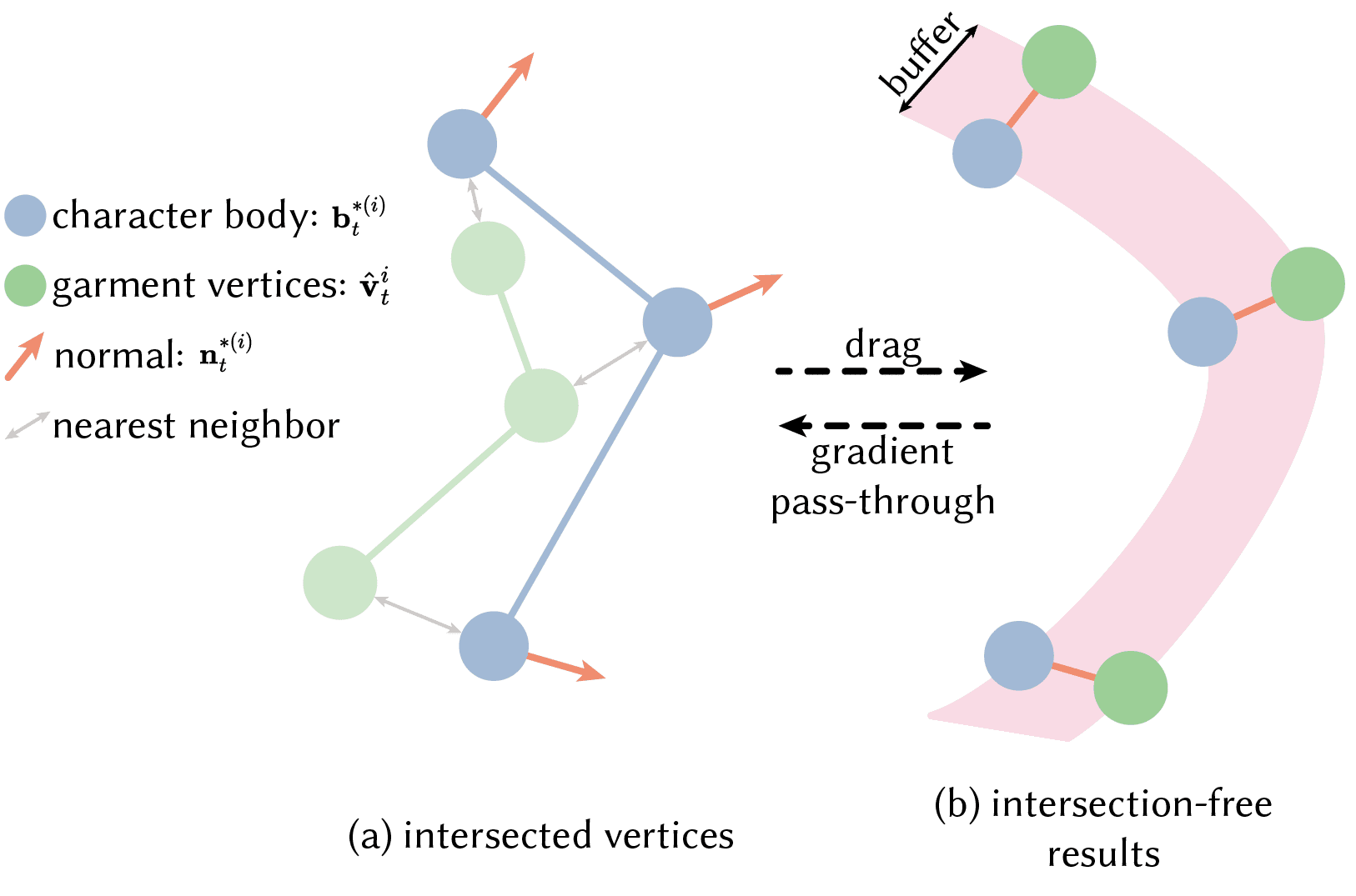}
  \end{center}
  \vspace{-20pt}
  \caption{\textbf{Plug-and-play intersection handling module.} (a) Intersections detected by the tangent plane of the nearest character body point. (b) Dragging intersected vertices out with a relaxation buffer.}
\label{fig:intersection_handling}
\vspace{-25pt}
\end{wrapfigure}

While our neural deformation fields are trained on physically correct data without intersections, garment-body intersections still occur occasionally on unseen motion sequences during inference. Therefore, we propose an intersection handling module that ``drags'' garment vertices intersected with the character's body outside. 

\vspace{-2mm}
\paragraph{\bf Determine Intersected Vertices.} For each vertex $\hat{\mathbf{v}}_t^i$ of the predicted garment deformation $\hat{\mathbf{G}}_t$, we first determine whether it is intersected with the character mesh $\mathbf{H}_t = \left\{\mathbf{b}_t^j \mid j\in\left\{1 \dots B\right\}\right\}\in\mathbb{R}^{B\times 3}$. We define the nearest character body point to $\hat{\mathbf{v}}_t^i$ as
\begin{equation}
    \mathbf{b}_t^{*\left(i\right)} = \operatorname{NN}(\hat{\mathbf{v}}_t^i)\text{,}
\label{eq:nn}
\end{equation}
where $\operatorname{NN(\cdot)}$ applies nearest neighbor searching. Suppose $\mathbf{b}_t^{*\left(i\right)}$ has vertex normal $\mathbf{n}_t^{*\left(i\right)}$, then we compute the signed projection norm of $\hat{\mathbf{v}}_t^i$ on the tangent plane of $\mathbf{n}_t^{*\left(i\right)}$ by an indicator function
\begin{equation}
    \mathbbm{1}(\hat{\mathbf{v}}_t^i) = \operatorname{ReLU}\left(\frac{\left(\hat{\mathbf{v}}_t^i - \mathbf{b}_t^{*\left(i\right)}\right)\cdot \mathbf{n}_t^{*\left(i\right)}}{\left\vert\left(\hat{\mathbf{v}}_t^i - \mathbf{b}_t^{*\left(i\right)}\right)\cdot \mathbf{n}_t^{*\left(i\right)}\right\vert}\right)\text{,}
\label{eq:inner_product}
\end{equation}
which indicates either the inner or the outer side $\mathbf{v}_t^i$ lies in. Thus, we can detect all the intersected vertices that are situated on the inner side of the tangent plane by checking the value of the indicator function of all the vertices.

\vspace{-2mm}
\paragraph{\bf Dragging.} Providing that $\mathbf{v}_t^i$ is detected as an intersected garment vertex, we aim to change its position from the inner side to the outer side of the tangent plane determined by $\mathbf{n}_t^{*\left(i\right)}$. Instead of optimizing a signed distance loss term, we directly ``drag" $\hat{\mathbf{v}}_t^i$ to its nearest character body point $\mathbf{b}_t^{*\left(i\right)}$, with an additional offset conforming to the direction of the body point normal $\mathbf{n}_t^{*\left(i\right)}$
\begin{equation}
    \hat{\mathbf{v}}_t^i = \hat{\mathbf{v}}_t^i + \operatorname{detach}\left(\mathbf{b}_t^{*\left(i\right)} + r\mathbf{n}_t^{*\left(i\right)} - \hat{\mathbf{v}}_t^i\right)\text{,}
\label{eq:drag}
\end{equation}
where $r$ is the relaxation coefficient that controls the buffer size, and $\operatorname{detach}(\cdot)$ operates gradient detaching to guarantee the ``dragging'' operation is differentiable.

\subsection{Real-time Clothed Character Animation with Motion Matching}
\label{sec:animation}

\begin{wrapfigure}{r}{0.42\textwidth}
\vspace{-35pt}
  \begin{center}
    \includegraphics[width=0.4\textwidth]{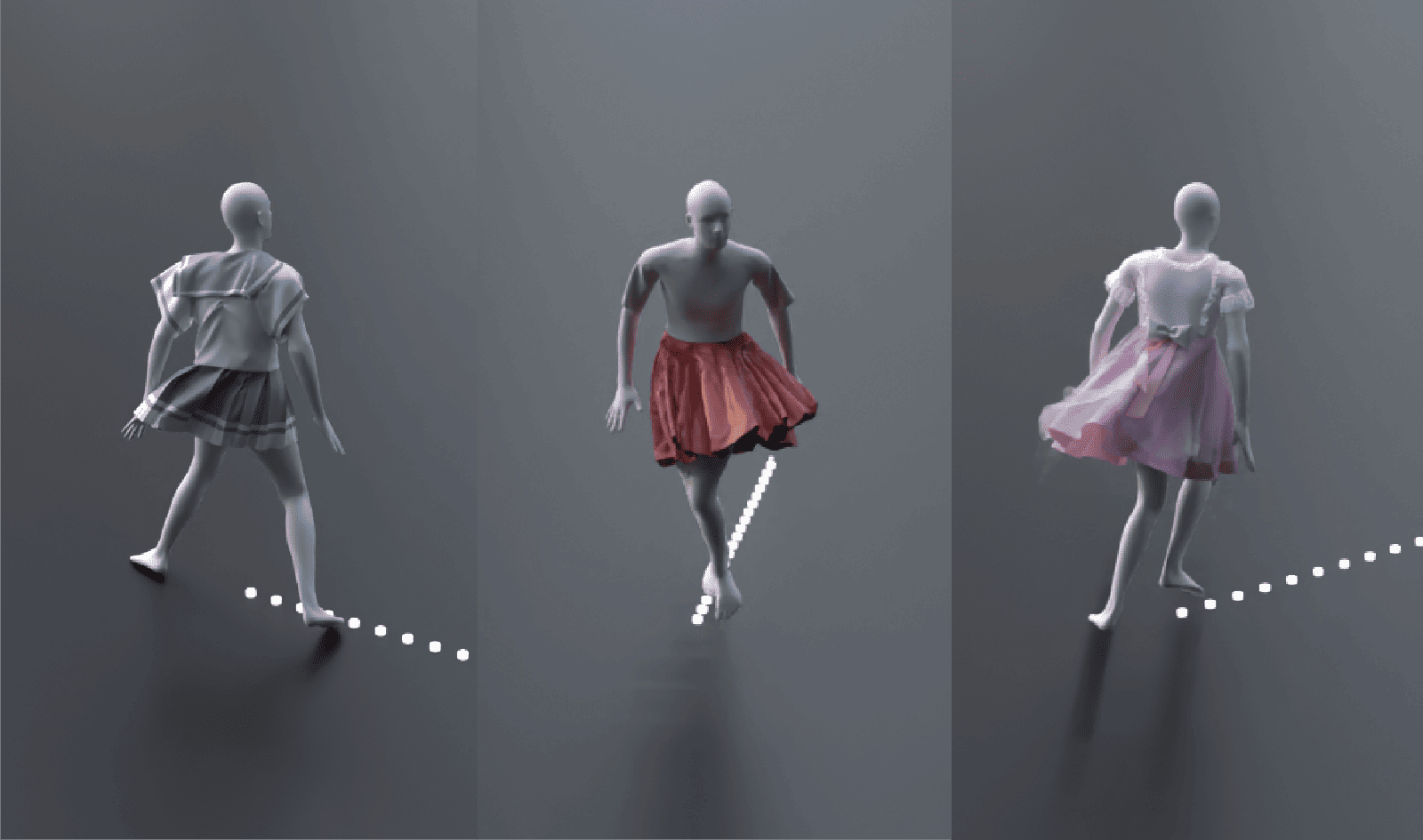}
  \end{center}
  \vspace{-20pt}
  \caption{\textbf{Real-time clothed character animation application.} We visualize character motion, future trajectory guidance, and inferred garment deformations to demonstrate the practical usage in downstream animation applications.}
\label{fig:animation}
\vspace{-30pt}
\end{wrapfigure}
We further develop a real-time demonstration of clothed character animation, leveraging the capabilities of UNIC to showcase its quality and efficiency. Specifically, as illustrated in Fig.~\ref{fig:animation}, we first enable interactive character animation using the state-of-the-art motion matching technique~\cite{starke2024categorical}, which allows users to control the character's trajectory, orientation, and locomotion in real time through mouse movements. Subsequently, the character motion obtained from motion matching is fed into UNIC, combined with a predetermined specific garment style as input. Our UNIC predicts garment deformation at the current timestamp and updates the geometry state in real time. This setup enables users to effortlessly animate clothed characters in a virtual environment using external control signals, jointly producing animations for both the character and the worn garment.

%% file: sections/4-exp.tex
\section{Experiments}
\label{sec:experiment}
\begin{figure*}[t!]
  \centering
  \includegraphics[width=\textwidth]{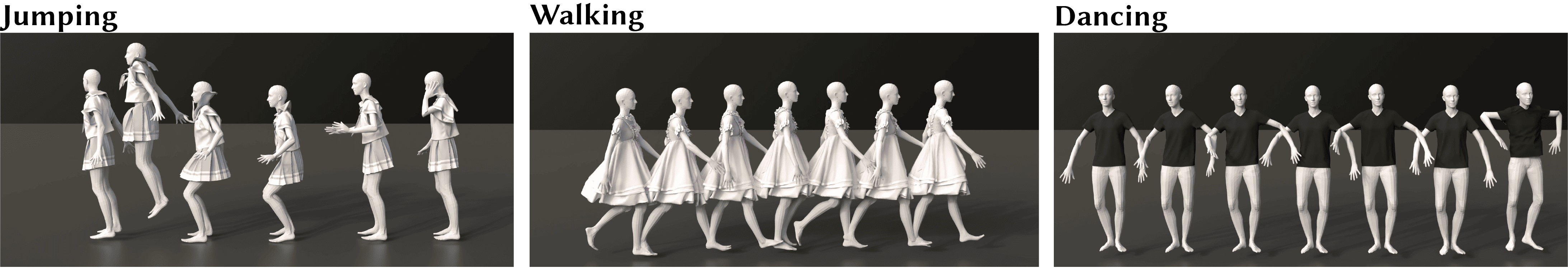}
  \vspace{-20pt}
  \caption{\textbf{Result Gallery} on testing set of CMU MoCap dataset.}
\vspace{-15pt}
\label{fig:cmu_test}
\end{figure*}

In this section, we first introduce our implementation details. Then, we present our results on CMU MoCap dataset (see Fig.~\ref{fig:cmu_test}) and compare our approach with several learning-based baselines on VTO-Dataset~\cite{santesteban2019learning}, VirtualBones-Dataset~\cite{pan2022virtualbones}, and our dataset. Subsequently, we conduct ablation studies to validate the effectiveness of our proposed model designs. And finally, we show the user study to demonstrate that our method is preferred in human feedback. We leave our data construction, user study details, and more qualitative results to our supplementary material.

\subsection{Implementation Details}
\label{sec:implemeentation_detail}
For our categorical character motion encoder, we set the channel and dimension of discrete latent space to $D_h=128$ and $D_w=8$ respectively. During training, we choose dropout probability as $0.25$. The hidden layer dimension of the motion encoder and deformation field is $4096$ and $256$, respectively. The relaxation buffer coefficient in the intersection handling module is set to $r=0.005$. The geometry training loss term is multiplied by $10^4$ for scaling the distance metric to centimeters. We utilize the AdamW optimizer \cite{loshchilov2017decoupled} with learning rate $\textrm{lr}=1\times10^{-4}$ and cosine annealing scheduler \cite{loshchilov2016sgdr}. The training batch size is adaptive to different garments due to various geometry complexities. Overall, we maintain that there are around $1.2\textrm{M}$ vertices in a single training batch. Our model is trained on 4 NVIDIA RTX 3090 GPUs in 300 epochs, which takes 2 hours for simple t-shirt and 20 hours for complex dresses. After training, our method generalizes to novel motion sequences in real time.

\begin{figure*}[t!]
  \centering
  \includegraphics[width=\linewidth]{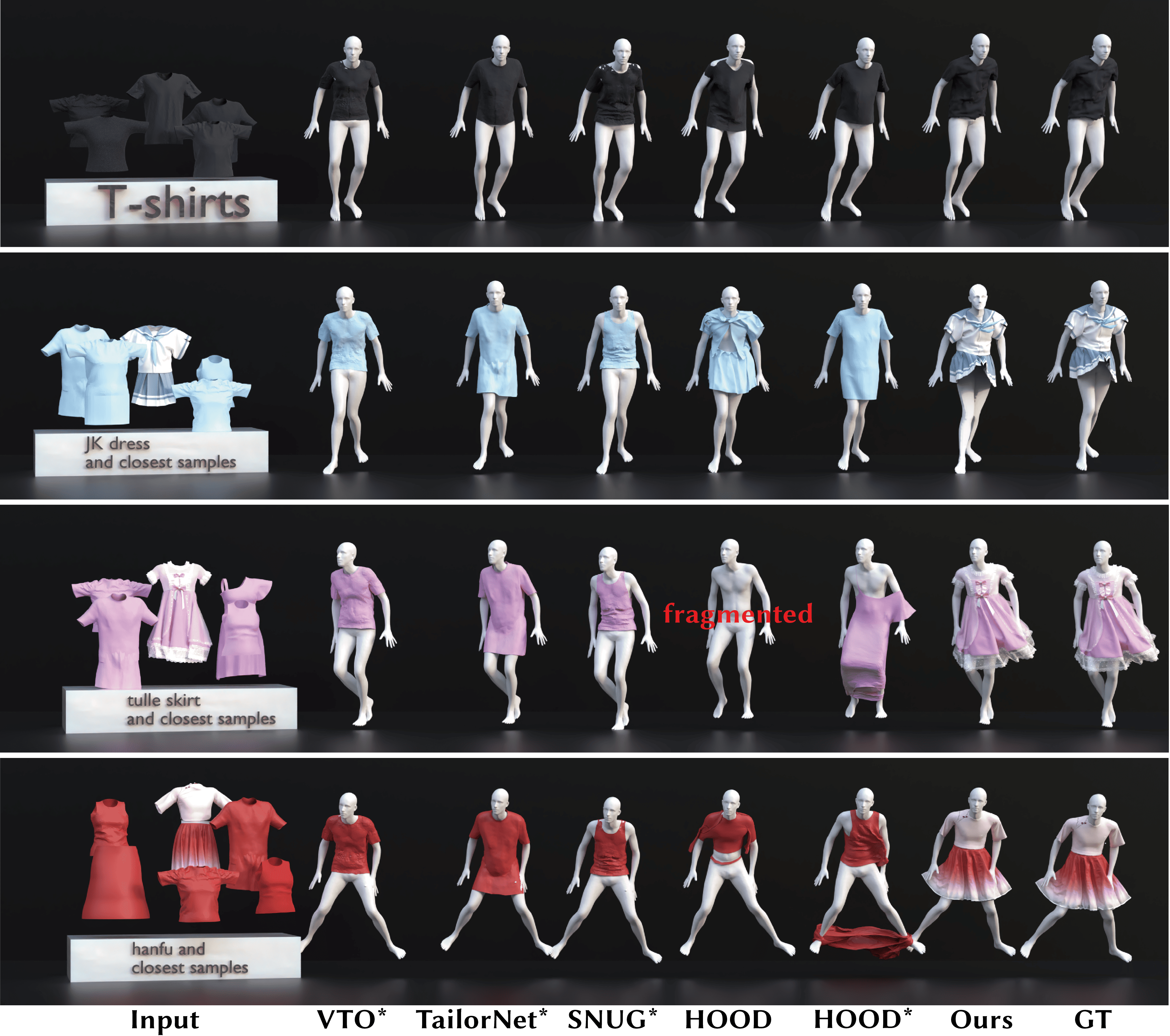}
  \vspace{-20pt}
  \caption{\textbf{Qualitative comparisons.} We present input garment geometries and simulation results of each method. Garment samples that are closest to our dataset within each method's data domain are selected to fairly evaluate the simulation quality, which is tagged with `*'.}
\vspace{-5pt}
\label{fig:comparison}
\end{figure*}

\subsection{Simulation Result Comparisons}
\label{sec:comparison}

\begin{table}[t!]
    \centering
    \scriptsize
    \setlength{\tabcolsep}{4.9pt}
    \begin{tabular}{lcccccccc}
        \toprule
        & \multicolumn{4}{c}{VTO-Dataset~\cite{santesteban2019learning}} & \multicolumn{4}{c}{VirtualBones-Dataset~\cite{pan2022virtualbones}}\\
        \cmidrule(lr){2-5}\cmidrule(lr){6-9}
        & \multicolumn{2}{c}{RMSE $\downarrow$} & \multicolumn{2}{c}{Hausdorff $\downarrow$} & \multicolumn{2}{c}{RMSE $\downarrow$} & \multicolumn{2}{c}{Hausdorff $\downarrow$}\\
        \cmidrule(lr){2-3}\cmidrule(lr){4-5}\cmidrule(lr){6-7}\cmidrule(lr){8-9}
        Method & T-shirt & Dress & T-shirt & Dress & Dress02 & Dress03 & Dress02 & Dress03\\
        \midrule 
        VTO~\cite{santesteban2019learning}          & 10.25 & 20.96 & 29.56 & 87.01 & - & - & - & -\\
        TailorNet~\cite{patel2020tailornet}    & 9.90  & 22.95 & 27.02 & 76.80 & - & - & - & -\\
        SNUG~\cite{santesteban2022snug}         & 12.30 & - & 34.91 & - & - & - & - & -\\
        HOOD~\cite{grigorev2023hood}         & 11.11 & 22.82 & 29.91 & 90.76 & - & - & - & -\\
        VirtualBones~\cite{pan2022virtualbones} & 10.52 & 19.91 & 31.51 & 83.39 & 25.10 & 24.50 & 95.76 & 65.99\\
        \cellcolor{lightgray!30}Ours         & \cellcolor{lightgray!30}\textbf{6.53} & \cellcolor{lightgray!30}\textbf{7.17} & \cellcolor{lightgray!30}\textbf{17.29} & \cellcolor{lightgray!30}\textbf{22.90} & \cellcolor{lightgray!30}\textbf{18.63} & \cellcolor{lightgray!30}\textbf{23.08} & \cellcolor{lightgray!30}\textbf{79.66} & \cellcolor{lightgray!30}\textbf{63.58}\\
        \bottomrule
    \end{tabular}
    \vspace{2pt}
    \caption{\textbf{Quantitative comparisons} with multiple baselines. We take two commonly used metrics to evaluate garment deformation accuracy on VTO-dataset~\cite{santesteban2019learning} and VirtualBones-dataset~\cite{pan2022virtualbones}, which are used in training by all the compared methods. `-' means the method has no corresponding model weights.}
\vspace{-20pt}
\label{tab:comparison}
\end{table}

\vspace{-2mm}
\paragraph{\bf Quantitative Comparisons} To quantitatively compare UNIC with baselines on commonly used benchmarks, we choose to evaluate on VTO-Dataset~\cite{santesteban2019learning} and VirtualBones-Dataset~\cite{pan2022virtualbones}, which are utilized for training in both UNIC and compared methods. We evaluate RMSE and Hausdorff distance in centimeters. As reported in Tab.~\ref{tab:comparison}, UNIC achieves consistently higher accuracy of garment deformation predictions, and surpasses state-of-the-art methods by $5\%\sim 60\%$ in RMSE and $3\%\sim 70\%$ in Hausdorff distance.

\vspace{-2mm}
\paragraph{\bf Qualitative Comparisons} We also conduct qualitative comparisons on complex garments in our dataset. However, since VTO~\cite{santesteban2019learning}, SNUG~\cite{santesteban2022snug}, and TailorNet~\cite{patel2020tailornet} only support limited garment variations of simple t-shirts and skirts, we sample garments from their data domains that most closely resemble the ones in our dataset for testing. For HOOD~\cite{grigorev2023hood}, we present results on our garments that were modified to satisfy its input requirement of sa ingle connected graph topology, as well as on its own garment samples that closely align with ours. As demonstrated in Fig.~\ref{fig:comparison}, UNIC is capable of simulating garments with more complex topologies and multiple sewing patterns. Additionally, unlike previous methods that rely on skinning weights and thus are restricted to tight-fitting garments, UNIC effectively deforms loose garments with elements including hemlines, ties, and cuffs. Furthermore, UNIC exhibits the ability to implicitly model various fabric materials, competently simulating deformations that conform to local physical attributes.

\begin{table}[t!]
    \centering
    \scriptsize
    \setlength{\tabcolsep}{17pt}
    \begin{tabular}{lcccc}
        \toprule
        & (\textasciitilde 3K) & (\textasciitilde 30K) & (\textasciitilde 59K) & (\textasciitilde 62K)\\
        Methods & T-shirt & JK Dress & Tulle Skirt & Hanfu\\
        \midrule 
        VTO$^*$~\cite{santesteban2019learning} & 2.1 & - & - & -\\
        TailorNet$^*$~\cite{patel2020tailornet} & 24.8  & - & - & -\\
        SNUG$^*$~\cite{santesteban2022snug} & 46.1 & - & - & -\\
        HOOD~\cite{grigorev2023hood} & 16.3 & 5.7 & 1.4 & 2.7\\
        MD CPU~\cite{md} & 28.4 & 3.0 & 1.3 & 1.6\\
        MD GPU~\cite{md} & 42.7 & 34.4 & 37.6 & 33.1\\
        \cellcolor{lightgray!30}Ours & \cellcolor{lightgray!30}\textbf{181.8} & \cellcolor{lightgray!30}\textbf{63.7} & \cellcolor{lightgray!30}\textbf{42.6} & \cellcolor{lightgray!30}\textbf{41.0}\\
        \bottomrule
    \end{tabular}
    \vspace{2pt}
    \caption{\textbf{Inference efficiency comparisons.} We test both learning-based methods~\cite{santesteban2019learning,patel2020tailornet,santesteban2022snug,grigorev2023hood} and physical simulation method~\cite{md,clo3d} on four garment styles in our dataset, ranging from a simple t-shirt to a complex tulle skirt. Given the same running environments and devices, we measure the running time of each method multiple times and determine the mean output fps. `*' represents that the tested garment geometry is slightly different from the given one in our dataset. `-' means that we cannot find an example in their datasets that is close enough to our geometries.}
\vspace{-15pt}
\label{tab:runtime}
\end{table}

\subsection{Runtime Cost Comparisons}
\label{sec:runtime}
In order to demonstrate the inference efficiency advantage of UNIC over learning-based methods~\cite{santesteban2019learning,patel2020tailornet,santesteban2022snug,grigorev2023hood} and professional software~\cite{md,clo3d}, we measure the runtime cost of each kind of garment style in our dataset by testing garment deformations on same motion sequences. Except for VTO~\cite{santesteban2019learning}, which cannot be deployed on GPUs, we test each method on a single NVIDIA RTX 3090 GPU 5 times using the testset of our dataset and take the mean as the overall runtime cost on each garment style. Notably, we report the cost of network forward operation only. However, the actual running cost of the GNN-based method~\cite{grigorev2023hood} is far more time-consuming due to the construction of the graph, which takes hours on complex garment topologies. We also compare with Marvelous Designer~\cite{md}, which is the ``gold standard'' in garment simulations, in both CPU and GPU-accelerated mode to show that UNIC achieves real-time inference performance while preserving physically realistic garment deformations. As shown in Tab.~\ref{tab:runtime}, UNIC consistently outperforms all the methods on garment topologies with the number of vertices ranging from $3\text{K}$ to $60\text{K}$. In particular, UNIC runs $1.5\sim 4$ times faster than industry-level software with GPU acceleration, yet in a totally differentiable and end-to-end manner.

\begin{figure}[t!]
    \centering
    \includegraphics[width=\linewidth]{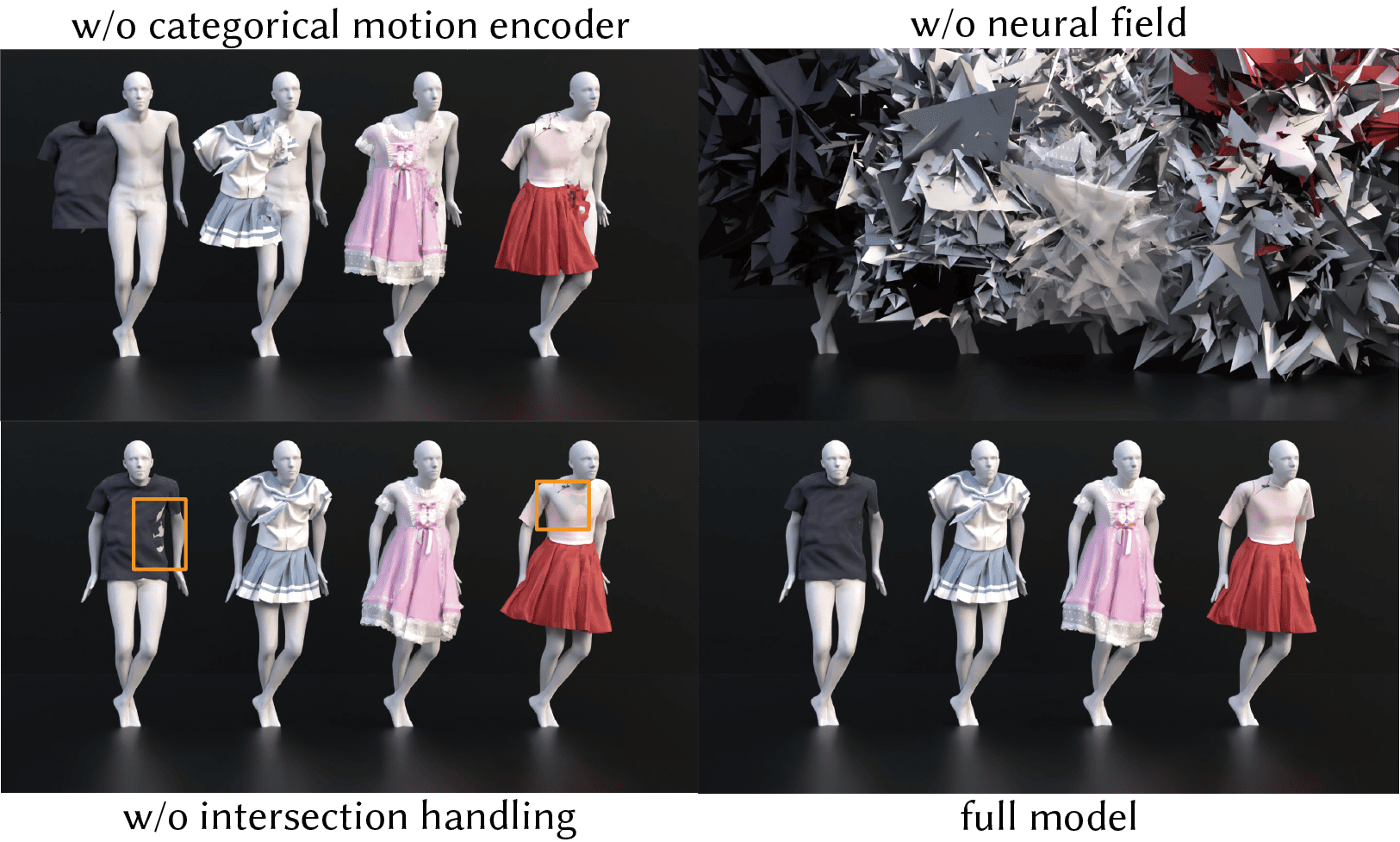}
    \vspace{-20pt}
    \caption{\textbf{Qualitative ablation study} on \textit{categorical motion encoder}, \textit{neural field}, and \textit{intersection handling module}.}
\label{fig:ablation}
\vspace{-5pt}
\end{figure}

\begin{table}[t!]
    \centering
    \scriptsize
    \setlength{\tabcolsep}{3.9pt}
    \begin{tabular}{lcccccc}
        \toprule
        & \multicolumn{3}{c}{RMSE $\downarrow$} & \multicolumn{3}{c}{Intersection Ratio $\downarrow$}\\
        \cmidrule(lr){2-4}\cmidrule(lr){5-7}
        Method & JK Dress & Tulle Skirt & Hanfu & JK Dress & Tulle Skirt & Hanfu\\
        \midrule 
        w/o categorical motion encoder & 42.21 & 46.61 & 53.04 & - & - & -\\
        w/o neural field & 447.13 & 397.37 & 443.67 & - & - & -\\
        w/o intersection handling & 9.33 & 7.83 & 13.81 & 35.93 & 21.17 & 35.21\\
        \cellcolor{lightgray!30}Ours & \cellcolor{lightgray!30}\textbf{9.72} & \cellcolor{lightgray!30}\textbf{8.67} & \cellcolor{lightgray!30}\textbf{13.62} & \cellcolor{lightgray!30}\textbf{0.00} & \cellcolor{lightgray!30}\textbf{0.06} & \cellcolor{lightgray!30}\textbf{0.14}\\
        \bottomrule
    \end{tabular}
    \vspace{2pt}
    \caption{\textbf{Quantitative ablation study} on our proposed \textit{categorical motion encoder}, \textit{neural field}, and \textit{interhection handling}. RMSE metric is used for evaluation, and Hausdorff distance is not included due to unacceptable computation cost on complex garments with numerous vertices in long sequences.}
\vspace{-25pt}
\label{tab:ablation_na}
\end{table}

\subsection{Ablation Study}
\label{sec:ablation}

To validate the effectiveness of each component of our method, we conduct three ablation studies on our \textit{categorical motion encoder}, \textit{neural field}, and \textit{intersection handling module}. The results are shown in Fig.~\ref{fig:ablation} and Tab.~\ref{tab:ablation_na}.

\vspace{-2mm}
\paragraph{\bf Evaluate Categorical Motion Encoder and Neural Field.} Firstly, we experiment with an alternative continuous motion manifold which directly maps two-frame character motions to a latent vector, without feature sampling on a learned distribution. As illustrated in Fig.~\ref{fig:ablation}, the continuous motion manifold can suffer from instant motion state transitions such as sprint and crash-stop, producing undesirable offsets due to the smooth latent space, and so cause distribution shifting in a roll-out inference manner. Secondly, we replace our neural field by using a global garment geometry feature to directly decode deformations. It is shown that the non-neural field method gives divergent results due to the limited capability of coarse-grain features, especially on complex garments with multiple patterns, multi-layer structure, and complicated topologies.

\vspace{-2mm}
\paragraph{\bf Evaluate Intersection Handling Module.} We further evaluate the proposed plug-and-play intersection handling module. It is illustrated that our intersection handling module can effectively eliminate garment-body mesh intersections (see Fig.~\ref{fig:ablation}), and significantly reduce the ratio of intersection chance by 20\%\textasciitilde30\% (see Tab.~\ref{tab:ablation_na}).

\begin{figure*}[t!]
  \centering
  \includegraphics[width=\textwidth]{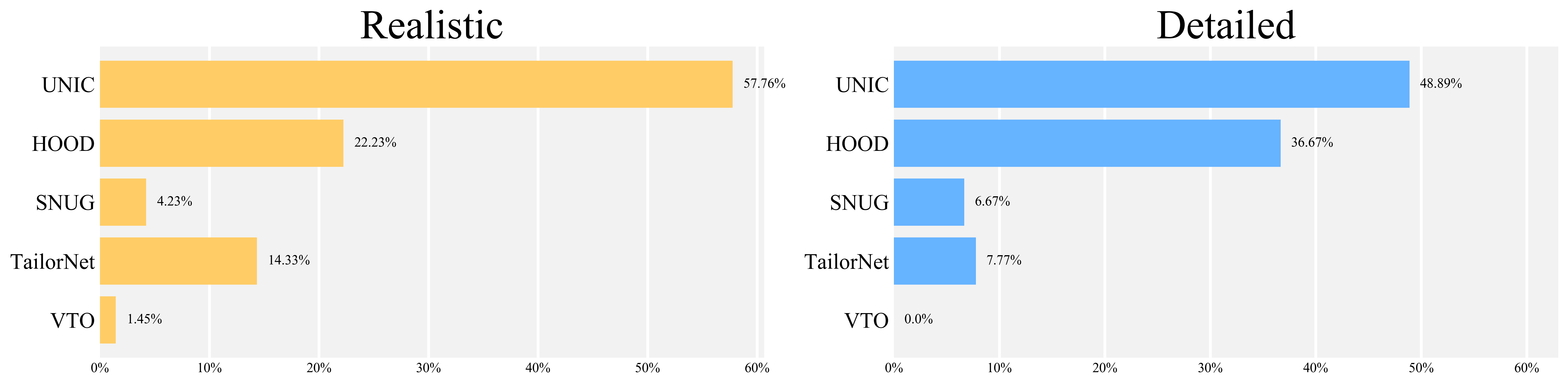}
  \captionof{figure}{\textbf{User study.} User studies of UNIC vs. other learning-based methods indicates strong preferences of UNIC in both simulation realism and detail.}
\vspace{-10px}
\label{fig:user_study}
\end{figure*}

\subsection{User Study}
\label{sec:user_study}
We conduct a comprehensive user study, structured around two primary evaluations: ``realism'' and ``detail'', which are aimed at measuring physical plausibility and capability of expressing fine-grained geometry deformation, respectively. We cut our test motion sequences into 8 clips around 20 seconds for result presentation. In total, more than 20 volunteers participated in our user study and answered questions to demonstrate their preference (see Fig.~\ref{fig:user_study}).

%% file: sections/5-conclusion.tex
\section{Conclusion}
\label{sec:conclusion}

We have presented UNIC, an instance-specific neural garment deformation field for simulating arbitrary garments on top of human character motions, in real time. Our method includes a categorical motion encoder, an MLP-based neural field on garment deformation space, and a plug-and-play intersection handling module. The key advantage of our method is the instance-specific neural field which can simulate arbitrarily complex garment deformations, consistent with the human character motions and the specific physical attributes of each garment sewing pattern. Querying this neural field concurrently on thousands of 3D garment vertices significantly enhances the inference efficiency of physical simulation on complex garments, and also mitigates the common problem of unrealistic artifacts caused by proximal skinning weights sampled from the parametric body model. We demonstrate that the garments deformed by our method are physically realistic, and readily usable in various applications such as gaming and animations.